\documentclass[pdflatex,sn-mathphys-num]{sn-jnl}

\usepackage{graphicx}%
\usepackage{multirow}%
\usepackage{amsmath,amssymb,amsfonts}%
\usepackage{amsthm}%
\usepackage{mathrsfs}%
\usepackage[title]{appendix}%
\usepackage{xcolor}%
\usepackage{textcomp}%
\usepackage{manyfoot}%
\usepackage{booktabs}%
\usepackage{algorithm}%
\usepackage{algorithmicx}%
\usepackage{algpseudocode}%
\usepackage{listings}%

\theoremstyle{thmstyleone}%
\theoremstyle{thmstyletwo}%

\theoremstyle{thmstylethree}%

\begin{document}

\title[Article Title]{Dyadic Partnership(DP): A Missing Link Towards Full Autonomy in Medical Robotics}


\author[1]{\fnm{Nassir} \sur{Navab}}
\author[2]{\fnm{Zhongliang} \sur{Jiang}}

\affil[1]{\orgdiv{Chair for Computer Aided Medical Procedures (CAMP)},
\orgdiv{School of computation, information and technology}, \orgname{TU Munich}, \city{Munich}, \country{Germany}}
\affil[2]{\orgdiv{Medical Intelligence and Robotic Cognition (MIRoC) lab}, \orgdiv{Mechanical Engineering Department}, \orgname{The University of Hong Kong}, \country{Hong Kong SAR, China}}



\abstract{For the past decades medical robotic solutions were mostly based on the concept of tele-manipulation. While their design was extremely intelligent, allowing for better access, improved dexterity, reduced tremor, and improved imaging, their intelligence was limited. They therefore left cognition and decision making to the surgeon. As medical robotics advances towards high-level autonomy, the scientific community needs to explore the required pathway towards partial and full autonomy. Here, we introduce the concept of Dyadic Partnership(DP), a new paradigm in which robots and clinicians engage in intelligent, expert interaction and collaboration. The Dyadic Partners would discuss and agree on decisions and actions during their dynamic and interactive collaboration relying also on intuitive advanced media using generative AI, such as a world model, and advanced multi-modal visualization. This article outlines the foundational components needed to enable such systems, including foundation models for clinical intelligence, multi-modal intent recognition, co-learning frameworks, advanced visualization, and explainable, trust-aware interaction. We further discuss key challenges such as data scarcity, lack of standardization, and ethical acceptance. Dyadic partnership is introduced and is positioned as a powerful yet achievable, acceptable milestone offering a promising pathway toward safer, more intuitive collaboration and a gradual transition to full autonomy across diverse clinical settings.}




\maketitle

\section{Introduction}\label{sec1}
Medical and surgical robotics are revolutionizing healthcare delivery in response to some of the most pressing societal challenges of our time~\cite{dupont2021decade,ciuti2025robotic}. With the global population aging rapidly and the burden of chronic diseases on the rise, healthcare systems are under growing pressure to provide high-quality, efficient, and consistent care. At the same time, billions of people worldwide still lack access to basic medical services, especially in remote or underserved regions~\cite{world2023tracking}.

Robotic technologies offer a compelling solution by extending the reach and capability of healthcare providers, enabling enhanced precision, control, and consistency across a wide range of clinical procedures. By integrating advanced sensing, artificial intelligence, and robotic actuation, these systems assist or automate tasks that traditionally rely heavily on human skill, such as minimally invasive surgery, ultrasound imaging, and targeted therapy delivery~\cite{jiang2023robotic,alterovitz2025medical, bi2024machine}. Their importance is growing rapidly due to increasing demands for higher accuracy, reduced recovery times, and broader accessibility to expert-level care. As healthcare continues to evolve toward personalized, data-driven, and minimally invasive interventions, medical robots are playing a central role in shaping the future of diagnostics and treatment~\cite{yip2025robot, schmidgall2025will, dupont2025grand, dupont2025medical}.

It is important to note that while earlier robotic systems primarily relied on teleoperation, the field is already shifting toward autonomy—driven by the need for intelligent systems that can operate reliably with minimal human intervention. The level of autonomy for medical robotics was first proposed by pioneers in the field in~\cite{yang2017medical}. Aattanasio~\emph{at al.} further discussed the autonomy level of the existing technology in the field of medical robotics~\cite{attanasio2021autonomy}. 
Teleoperated systems are fundamentally constrained by the skill, attention, and availability of human operators. In contrast, full autonomy enables robots not only to replicate expert tasks but to optimize and execute them with superhuman consistency and scalability. This paradigm shift opens the door to more standardized care, continuous operation in high-demand settings, and the potential to deliver expert-level procedures where human expertise is scarce or absent.

Shared autonomy is an existing concept that allows robots to provide semi-autonomous assistance, such as maintaining optimal probe contact during robotic ultrasound while the clinician navigates the scan path~\cite{huang2024robot}, or enforcing safety constraints in orthopedic surgery through virtual fixtures~\cite{li2007spatial}. These applications demonstrate how shared control can enhance precision, reduce cognitive load, and build trust. However, shared autonomy is loosely defined and limited in scope. Most implementations focus narrowly on task-level role allocation—such as assigning specific degrees of freedom or predefined subtasks—without robust mechanisms for dynamic collaboration, smooth transitions of control, or real-time intent understanding. Critically, there is no widely accepted framework for how autonomy should adapt to varying task complexity, uncertainty, or user confidence. As a result, shared autonomy systems are often engineered in an ad hoc manner, limiting their generalizability, scalability, and effectiveness in real-world clinical environments.


With the increase of the intelligence of the surgical systems and the fast development of the surgical data science, the medical robotics needs to go way beyond the limitations of current shared autonomy frameworks. The idea is no more about shared control but also shared dynamic data analysis and decision making. This needs novel concepts and paradigms taking the full capabilities of AI-enabled surgical robotic systems into account. Here, we therefore introduce the new concept of Dyadic Partnership (DP). 
It is important to note that the majority of the current robotic systems are mostly aiming at enhancing human capabilities. It is no surprise that these systems usually rely on one or two tele-manipulated arms and focus on enhancing human visual access to the surgical scene. This also results in limiting the perspectives and the basic design of the shared control concepts. However, future clinical robots could have many more arms and actuators controlled through different physics and mechanics, and many more sensors for dynamic observation and monitoring of the targeted anatomy and physiology, including co-registered dynamic iOCT, robotic Ultrasound, Optoacoustics, X-ray and other interventional modalities. As the sensing and interaction capabilities of clinical robots increase, the role of efficient and intelligent human/machine communication becomes one of the main keys for acquiring the trust of the surgeons and giving them enough tools to fully master the process in order for them to accept to take the responsibility of the automatic execution of surgical tasks by partially or fully automated robots. In order to get to such objectives a major requirement is a high-level dynamic discussion and partnership between the surgeons and the robots. We call this new paradigm the Dyadic Partnership(DP), in which both parties respect each others knowledge and capabilities and discuss their plans and actions with the focus on the patient outcome.

\begin{figure}[h]\label{fig1}
\centering
\includegraphics[width=0.9\textwidth]{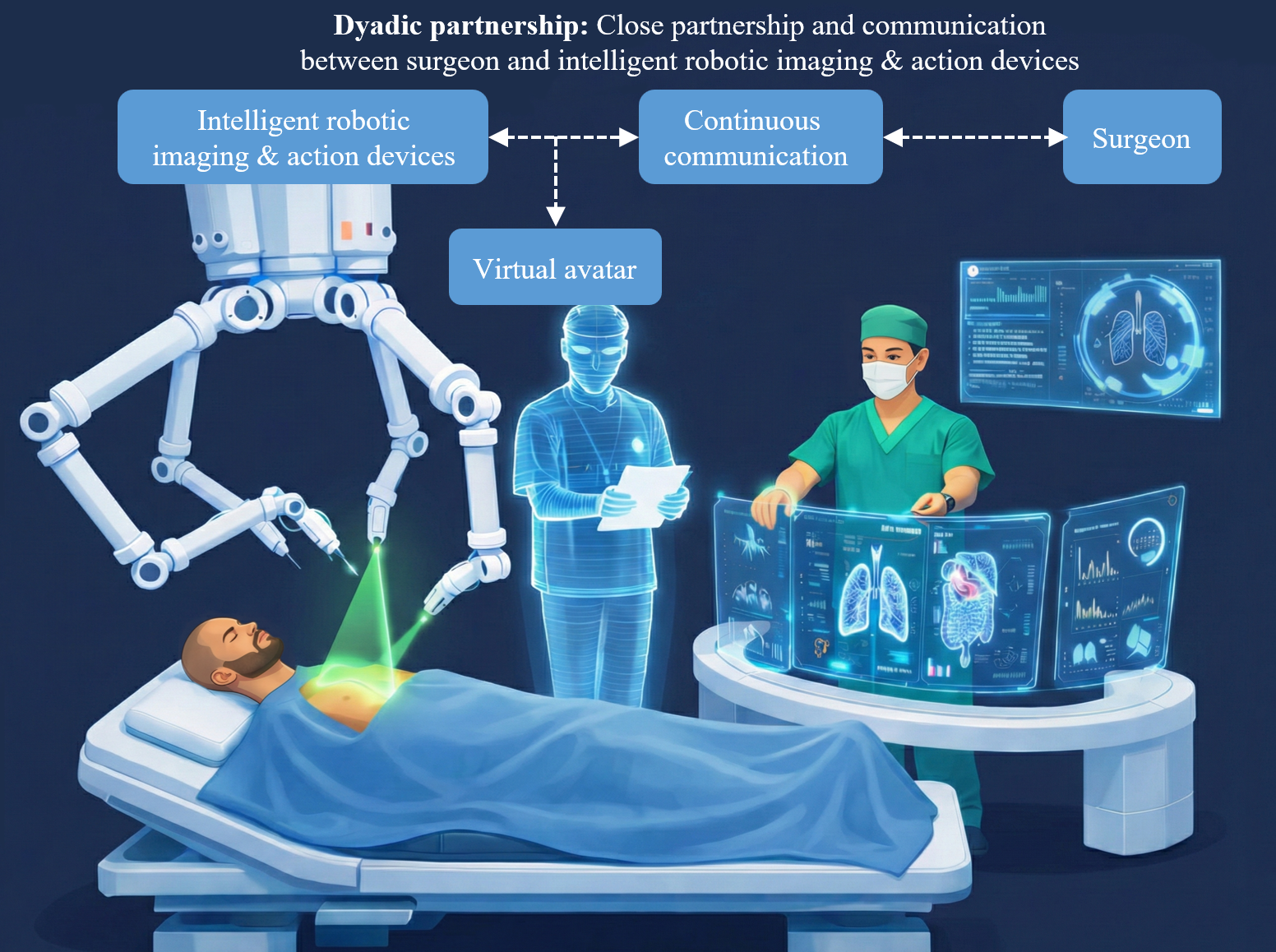}
\caption{Dyadic Partnership between the surgeon and intelligent autonomous robotic imaging and action devices supports informed trust in the technology and strengthens the physician’s ownership of clinical decision-making and patient outcomes.
}
\end{figure}

DP refers to a collaborative and structured framework in which a human and a robotic or intelligent agent engage in a dynamic, bidirectional relationship—sharing knowledge through interactive communication and intuitive perceptual visualization to enable transparent, reliable decision-making in pursuit of a common clinical goal (see Fig.~\ref{fig1}). Dyadic partnership offers a clearer definition that emphasizes mutual respect, real-time adaptation, and contextual awareness. Unlike traditional models that view robots as passive tools or task executors, and the fully automatic systems that expect surgeons to bear the risk for such automatic actions,  DP positions the robot and the surgeon as intelligent collaborators capable of proactive support, adaptive interaction, and mutual understanding throughout the clinical workflow. This framework enables intelligent agents to interpret human intent, adapt to changing conditions, and even proactively initiate supportive actions, while enabling the surgeon to fully observe and evaluate the perception, decision, and proposed actions of the robotic systems. DP fills a critical gap in current autonomy research by offering a more holistic, interactive approach that aligns with the complex, multi-agent nature of real-world healthcare environments and the complex process of multi-modal sensing, dynamic decision making, coordinated real-time planning, risk analysis resulted in trusted actions. 

\section{Benefits of Dyadic Partnership}
The sense–think–act cycle is central to robotic decision-making and behavior, but its implementation varies across different autonomy paradigms~\cite{dupont2025grand}. In teleoperation, the robot primarily serves as an extension of the human operator. It senses the environment and relays information—such as visual or force feedback, while the human handles all reasoning and decision-making. The robot’s actions are direct responses to human input, with minimal independent autonomy. In contrast, classical autonomy uses structured models and rule-based planning to interpret sensor data, reason about tasks, and execute actions independently. While effective in controlled settings, these systems often lack the flexibility to adapt to real-world uncertainty or nuanced human behavior.

\par
Learned autonomy introduces a shift toward data-driven, adaptive decision-making. Instead of relying on hand-crafted models, robots trained via machine learning learn to map raw sensory inputs (e.g., video, ultrasound) directly to actions or decisions based on demonstrations or large datasets. This allows for greater generalization and performance in unstructured environments, but it can introduce issues related to interpretability and safety. Across these paradigms, the sense–think–act cycle moves from being human-dominated (teleoperation) to robot-dominated (learned autonomy) under human supervision. However, this transition still lacks mechanisms for real-time, collaborative interaction—highlighting the need for frameworks like Dyadic Partnership, which tightly integrates both human and robot contributions throughout the entire cycle.

\par
Integrating Dyadic Partnership into the classical sense–think–act paradigm reframes the robot not as a passive executor, but as an interactive collaborator. In the sensing phase, the system must go beyond human and environmental perception and one side sense many information which human cannot capture directly through his or her limited sensing capabilities but also capture human intent, physiological cues, and task context, which relate to human perception and decision making process. Advanced perceptual visualization becomes necessary for the AI-embodied robotic system to provide information to surgeon making the intention of the robot clear to its surgeon partner. This will allow the partners during thinking to jointly reason about both the clinical goal and the partner’s behavior—adapting strategies, anticipating needs, discussing tactical and strategical actions, and/or negotiating control. In the acting phase, the robot executes motions or decisions that are informed not only by algorithmic output but also by the evolving human-robot relationship. This continuous feedback loop enables real-time, adaptive collaboration—making the robot capable of assisting, correcting, or even initiating actions in a clinically meaningful way. Thus, Dyadic Partnership transforms the traditional cycle into a shared cognitive and physical loop, grounded in mutual awareness and co-adaptation. Due to their scalable nature and capacity for holistic understanding of surgical procedures, such intelligent robot-surgeon partnership can significantly accelerate the path toward full autonomy in medical robotics.

\subsection{Improved Human-Robot Understanding and Trust}
A key benefit of Dyadic Partnership is the advancement of mutual understanding between human clinicians and robotic systems, which forms the foundation for building trust in high-stakes clinical environments. Traditional robotic systems often operate as passive tools or black-box agents, offering little transparency into their decision-making processes or adaptability to human intent. In contrast, Dyadic systems could also be designed to perceive and interpret human behavior—such as motion patterns, gaze direction, force input, or verbal cues—and adjust their actions accordingly. This enables the robot to respond in ways that considers the required alignment with the clinician's goals and expectations, reducing the cognitive burden of micromanagement and allowing more natural, intuitive interactions.

This enhanced responsiveness fosters trust through predictability, transparency, and shared situational awareness. When clinicians see that the robot can reliably interpret their intent, adapt in real time, and handle routine tasks with minimal intervention, they become more confident in its use. Over time, this trust enables a deeper collaboration where clinicians are willing to delegate higher-level tasks, paving the way for more advanced autonomous behaviors. Importantly, trust is not built through technical accuracy alone—it arises from the system's ability to communicate, to fail gracefully, and to recover in ways that maintain human confidence. Dyadic Partnership addresses these challenges by embedding mechanisms for active communication and adaptive behavior throughout the entire sense–think–communicate-act  cycle. It is of course sometimes important to act fast but considering keeping the partner in the loop as soon as possible would make the system a strong, communicative and reliable partner. 

\subsection{Reduced Cognitive and Physical Workload}
Dyadic Partnership also significantly reduces cognitive and physical workload for clinical professionals by fostering an apprenticeship-inspired model of shared learning and reciprocal support.
In the early stages of this relationship, the human expert assumes the role of a teacher or supervisor, guiding the robot through demonstrations, corrections, and high-level decisions. The intelligent agent, much like a trainee, learns not only task execution but also the underlying context, intent, and variability inherent in real-world clinical workflows. This human-in-the-loop learning phase allows the system to develop nuanced capabilities while ensuring safety and transparency—critical in high-risk environments, such as surgery or image-guided intervention.

As the intelligent system matures and acquires a holistic understanding across diverse procedures, patients' anatomy and physiology, the dynamic may shift—the robot gradually transitions from student to assistant, and in some cases, to mentor or guide. In repetitive or routine scenarios, the robot can proactively suggest strategies, optimize workflow steps, or take over subtasks entirely, freeing the human to focus on higher-level clinical decision-making. This role reversal reduces mental fatigue, minimizes manual effort, and enhances procedural efficiency. Importantly, this evolving partnership does not displace the human expert but redefines the collaboration—creating a flexible, adaptive support system where the robot continuously offloads routine burdens while remaining responsive to human oversight and contextual input.

\subsection{Scalability and Knowledge Transfer}
One of the most transformative benefits of Dyadic Partnership is its potential to enable scalability and knowledge transfer across healthcare systems. Unlike conventional automation, which is often rigid and domain-specific, the proposed dyadic systems are designed to learn from expert interactions in context-rich environments. As robots observe and adapt to the behaviors, strategies, and decision-making patterns of experienced clinicians, they build internal representations that go beyond rote execution. These learned models—grounded in both sensor data and collaborative intent—can be transferred across different clinical tasks, anatomical regions, or even specialties. This capacity allows robotic systems to generalize expert-level insights to new environments, making them especially valuable in settings with limited access to skilled personnel.

Moreover, once an intelligent agent has acquired a robust understanding of certain procedures through dyadic interaction, this knowledge can be scaled and shared across platforms and institutions. For example, a robotic assistant trained in ultrasound-guided liver interventions at a central hospital can inform systems deployed in rural clinics, enabling them to deliver expert-level performance without requiring extensive local training. Similarly, a system that has learned safety constraints or best practices from one operator can adapt those principles to assist others with different techniques or preferences. In this way, Dyadic Partnership not only enhances individual performance but also becomes a vehicle for democratizing clinical expertise, improving the quality and consistency of care across geographical and institutional boundaries.

\subsection{Adaptability to Clinical Complexity and Uncertainty}
Clinical environments are inherently complex, dynamic, and uncertain, with frequent variability in patient anatomy, disease presentation, and procedural conditions. Traditional autonomous systems often struggle in such settings due to their reliance on rigid models or predefined workflows that assume ideal, repeatable conditions. Dyadic Partnership addresses this limitation by enabling robots to adapt fluidly and contextually through continuous interaction with human experts. Rather than executing fixed plans, the robot can adjust its behavior based on real-time cues from the human partner—such as changes in force, motion patterns, or verbal requests—and respond to unanticipated events collaboratively. This adaptability allows the system to function robustly even in incomplete or noisy data conditions, making it more reliable for real-world deployment.

In practice, this means a dyadic system can, for instance, assist in navigating a complex anatomical region with variable tissue stiffness, accommodate intraoperative changes such as patient movement or tool slippage, or defer control when human judgment becomes critical. Importantly, the adaptability is not unidirectional; the robot can also proactively inform or guide the human when its confidence is high or when deviations from the expected pattern are detected. This bidirectional flexibility transforms the robotic agent from a passive executor into a collaborative problem solver, capable of managing uncertainty in ways that traditional automation cannot. In doing so, Dyadic Partnership becomes a key enabler for safe and effective deployment of intelligent systems in unpredictable and high-stakes clinical environments.

\subsection{Accelerated Pathway to Full Autonomy}
Dyadic Partnership serves as a practical and strategic bridge toward full autonomy in medical robotics. By enabling gradual, real-time collaboration between humans and intelligent agents, it allows robotic systems to learn from expert behavior, adapt safely within clinical workflows, and incrementally take on greater responsibility. This progressive transition builds user trust, ensures clinical relevance, and supports regulatory acceptance—collectively accelerating the development and deployment of fully autonomous systems in healthcare. Moreover, with the integration of a shared world model~\cite{lecun2022path, ha2018world}, dyadic systems can intelligently visualize upcoming actions, predicted outcomes, and regions of uncertainty. For example, the robot may proactively highlight specific imaging planes, anatomical structures, or decision points that warrant clinician input—facilitating a converged, well-justified clinical strategy. Such transparent, interactive decision-making not only enhances collaboration but also provides a clear audit trail of how decisions are made, paving the way for regulatory clearance and clinical adoption by aligning with safety, accountability, and explainability standards.

\section{Enabling Dyadic Partnership in Medical Robotics}

\subsection{Intelligent Presentation of Multi-Modality Sensing}
Clinical decision-making relies on a deep understanding of sub-surface anatomies and the dynamic behavior of living tissues—expertise honed through years of surgical training. To establish a truly equal and collaborative dyadic partnership between surgeons and intelligent robotic systems, robots must not only acquire comparable anatomical and physiological understanding but also communicate their internal reasoning in an interpretable manner. This requires demonstrating comprehension through multiple sensing modalities—visual, tactile, acoustic, and force-based—so that trust, transparency, and bidirectional dialogue can emerge. A useful analogy is autonomous parking: modern systems do more than compute trajectories, where they visualise the inferred environment, including views beyond human perception and predicted motion paths, enabling users to assess the system’s mental model. Surgical robots must provide a similar level of intelligible situational awareness, but under far greater complexity. 

Visualisation remains the most intuitive mechanism for presenting robotic understanding. Augmented and extended reality can expose inferred anatomical layers, tool–tissue interactions, or predicted tissue motion based on perceptual visualisation~\cite{bichlmeier2007contextual, lerotic2007pq}. Because surgery involves extensive reasoning about invisible structures, robots must continuously register real-time sensor data with preoperative imaging or anatomical atlases to achieve precise or semantic alignment, respectively~\cite{wang2024recursive,jiang2024class}. Such registration enables perception beyond the visible surface, revealing sub-surface organs, risk zones, and safe navigation corridors. In a dyadic partnership, however, visualization must extend beyond showing the ``next action” of a trained agent. Instead, it should support dynamic strategy negotiation: presenting alternative plans, expected outcomes, and uncertainty estimates, much like a surgeon adapting to intraoperative variations or unexpected bleeding. This level of transparency allows human and robotic agents to converge on situationally adaptive decisions. Recently, Chen~\emph{et al.} evaluated state-of-the-art video generation models (i.e., world models) in surgical settings and revealed a substantial gap: although these models can reproduce surgical scenes with impressive photorealism, they lack a true understanding of surgical practice, failing to respect fundamental principles of action, consequence, and procedural strategy~\cite{chen2025far}.

Importantly, communication should not rely solely on vision. Acoustic cues, including real and generated~\cite{matinfar2025tissue}, can convey tool–tissue dynamics, while emerging language-based interfaces—powered by vision–language models—enable natural, conversational interrogation of system intent and confidence. Tactile and force signatures further enrich this communication channel~\cite{zhang2025tactile, lee2020nanomesh}, reflecting tissue stiffness, structural boundaries, or imminent risk. By creatively integrating these modalities, future medical robots can express a multisensory understanding of the operative field, approaching the perceptual richness of human surgeons and enabling a deeper form of collaborative intelligence.

\subsection{Foundation Models for Clinical Intelligence}
Foundation models represent a transformative step toward enabling dyadic partnership by serving as large-scale, pre-trained systems that encode generalizable knowledge across procedural, anatomical, and task domains. These models, often built on massive amounts of multimodal clinical data—such as images, text, videos, and sensor streams—form a shared, reusable backbone for downstream tasks in medical robotics. Unlike traditional systems trained on narrow, task-specific datasets, foundation models can support reasoning, intent prediction, and semantic understanding across a wide range of clinical procedures and settings.

In medical imaging, recent vision-language models (VLMs) such as BioViL~\cite{bannur2023learning}, MedCLIP~\cite{wang2022medclip}, and GLoRIA~\cite{huang2021gloria} have shown the ability to learn joint embeddings between radiology images and text reports, enabling zero-shot disease classification, report generation, and image-text retrieval. These models can contribute to the perceptual core for robotic systems by enabling rich semantic understanding of visual input—for example, identifying anatomical landmarks or pathology descriptions based on free-text queries.

In the context of medical robotics, a natural extension is vision-language-action (VLA) models, which integrate perceptual understanding with action planning and execution. Early examples include models that learn to follow natural language instructions in simulation or robot-assisted environments (e.g., PaLM-E~\cite{driess2023palm} or RT-2~\cite{zitkovich2023rt}) and those fine-tuned on procedural demonstrations in surgical settings. In future applications, a VLA model in robotic ultrasound could, for instance, interpret a spoken command like ``find the liver and scan along the intercostal space", understand the semantic goal, localize the relevant anatomy, and generate an adaptive motion plan—all grounded in prior procedural knowledge. By leveraging these foundation models, intelligent agents can generalize across tasks, adapt to new users, and support more fluid, semantically rich interaction with human collaborators—making them a cornerstone for building Dyadic Partnership in real-world clinical robotics.

\subsection{Multimodal Intent Recognition and Context Understanding}
For a medical robot to function as a true Dyadic Partner, it must not only perceive the physical environment but also understand the intentions, goals, and contextual cues of its human collaborators. Achieving this requires the fusion of multiple sensory modalities, such as visual input, speech, gaze patterns, tool dynamics, and physiological signals, to develop a comprehensive, real-time understanding of both the task at hand and the human user. While humans acquire clinical skills through experience and formal education, machines have the potential to go beyond human capabilities by integrating rich, high-dimensional sensor data. For instance, autonomous vehicles are equipped with cameras, LiDAR and radar to provide a broader and more detailed view of their surroundings than a human driver could perceive. In autonomous cars, this information is visualized or sonified within dynamic user interfaces to make it perceptible to the human user. Similarly, in a Dyadic Partnership, medical robots are not only expected to process dynamic multimodal information but also to share relevant insights, reasoning, or contextual feedback with their human partners—building more transparent, informed, and collaborative decision-making. This brings in the need for advanced perceptual user interfaces that clearly summarize the results of dynamic sensor fusion and predictive multimodal world model generation and provide adaptive perceptual feedback to the Dyadic Partners. 

Multimodal intent recognition involves capturing subtle cues that reveal what the clinician is trying to achieve and how they are interacting with the robot and the patient. For example, eye gaze and head pose can signal the surgeon's focus of attention or upcoming action; tool trajectories and motion patterns can indicate task phases or precision requirements; voice commands or natural language cues can convey high-level goals or requests for assistance; and physiological signals like heart rate variability may reflect cognitive load or stress, useful for adaptive robot behavior. At the same time, the surgeon also needs to be aware of the internal status and intention of the robot when it aims at performing fully or partially automated actions. In fact, both Dyadic Partners need to observe the intention but also the trust of the partner in every decision and action. Surgery is about optimized decision and calculated risk taking. Therefore, both partners need to be fully aware of each other's plans as well as certainty in the prediction of the outcome. 

In practice, consider a robotic ultrasound assistant that combines gaze tracking~\cite{bi2025gaze,men2023gaze} and probe motion to predict when the clinician is about to adjust imaging depth or shift the scanning plane—allowing the system to proactively stabilize the probe or optimize image parameters. In a robotic surgery scenario~\cite{yip2023artificial, saeidi2022autonomous,long2025surgical}, the robot could fuse haptic feedback, instrument motion, and verbal cues to anticipate a tissue dissection step and adjust the tool's compliance accordingly. Similarly, in rehabilitation robotics~\cite{luo2024experiment}, interpreting both verbal encouragement from a therapist and muscle activation from a patient can help the system dynamically modulate support.

Achieving robust multimodal fusion requires addressing challenges such as temporal synchronization, cross-modal uncertainty, and context dependency. However, when done effectively, it enables the robot to shift from passive sensing to interactive perception, where observations are tied directly to collaborative action and real-time decision-making—laying the foundation for more intelligent and responsive dyadic systems in healthcare. Please note that the surgeon also needs to observe similar parameters including attention, movement and force, when the robot is doing an action in order to coordinate their actions. It is therefore important that actions and intentions of the robotic systems are displayed to their dyadic partners through human perceptual channels of vision, auditory and haptic.

\subsection{Human-Robot Co-Learning and Personalization}
In a true Dyadic Partnership, learning must occur in both directions—the robot continuously adapts to the clinician, while the clinician also learns to trust and leverage the robot's capabilities. This two-way process, known as human-robot co-learning, enables personalized, evolving collaboration that becomes more efficient, intuitive, and robust over time.

From the robot’s perspective, co-learning involves building models of individual user preferences, skill levels, and task-specific styles. For example, a robotic assistant may learn how a particular surgeon prefers to approach liver resections—whether they favor blunt dissection or sharp cutting—and then adapt motion planning, force thresholds, or visual feedback accordingly. This personalization allows the robot to provide support that feels natural and non-intrusive, improving efficiency while reducing friction in the interaction.

Conversely, the human also benefits by learning from the robot, especially as intelligent systems begin to accumulate knowledge across users and institutions. For instance, a novice operator may receive real-time guidance during robotic ultrasound scanning—such as highlighting anatomical landmarks, suggesting probe trajectories, or flagging missing coverage—based on prior expert demonstrations. Over time, the human operator can rely less on external instruction and more on the robot as a context-aware mentoring partner, ultimately improving clinical competency and confidence.

This co-adaptive relationship also fosters mutual trust: the clinician learns the robot’s behavioral patterns and limitations, while the robot refines its model of the human’s intent and expertise~\cite{jiang2024intelligent}. This is especially important in safety-critical environments where miscommunication or rigid behavior can lead to clinical error. By enabling personalized learning loops, co-learning supports scalable deployment across diverse users and institutions—laying the groundwork for intelligent systems that integrate seamlessly into individualized clinical workflows.

\subsection{Explainable and Trust-Aware Interaction Models}
For dyadic partnership to be safe, effective, and widely accepted in clinical settings, intelligent systems must not only perform well—they must also be understood and trusted by their human collaborators. This requires the development of explainable and trust-aware interaction models, which transparently communicate the robot’s goals, reasoning process, and uncertainties in real time.

A key aspect of this is intuitive visualization. Clinicians often operate under time pressure and cognitive load, so the robot must present its internal state—such as target intentions, predicted outcomes, or confidence levels—through clear, actionable visuals. For example, in robotic ultrasound, the system could overlay real-time confidence maps or planned probe trajectories directly on the imaging screen. In surgical robotics, projected visual cues or AR-based indicators could show the robot’s intended path or safety boundaries. These visualizations reduce ambiguity and allow clinicians to quickly assess whether to accept, correct, or override the robot’s decisions.

In parallel, intuitive communication mechanisms—such as auditory feedback, gesture-based cues, or haptic signals—can support smooth interaction without interrupting clinical flow. For instance, a robot might verbally alert the user when it detects unusual tissue stiffness, or respond to a simple hand motion to pause or shift its assistance mode. Instead of the verbal feedback the robot may also use alternative auditory feedback~\cite{matinfar2025tissue, ha2025full}, which could be more subtle and pleasant, if it is designed by medical sonification experts. These communication pathways are critical for maintaining shared situational awareness, preventing silent errors, and building calibrated trust over time. By combining intuitive visual and auditory feedback with models of human trust dynamics, explainable interaction becomes not just a feature—but a foundational requirement for safe, collaborative autonomy in real-world clinical practice.

\section{Challenges}
\subsection{Scarcity of Medical Data}
Data scarcity remains one of the most significant barriers to progress in medical AI and robotics—particularly in developing intelligent agents for Dyadic Partnership, which requires rich, diverse, and context-aware datasets. Unlike consumer applications where massive datasets are readily available, medical data collection is inherently constrained by privacy regulations, institutional silos, and clinical logistics. This challenge becomes even more pronounced when collecting high-quality VLA datasets essential for enabling autonomous behavior. Beyond data collection, the organization and standardization of large-scale, multimodal robotic datasets present additional hurdles. There have however been some efforts to address this in different clincial settings. Chen~\emph{et al.} recently introduced Robo-DM—an efficient, open-source, cloud-based data management toolkit designed for collecting, sharing, and learning from robot data, which supports synchronized video, textual inputs, and streams from multiple cameras and sensors, offering a scalable solution to streamline dataset handling in complex robotic systems~\cite{chen2025robo}. 

First, data collection itself is challenging. Capturing high-quality, multimodal clinical data—such as video, ultrasound, iOCT, Optoacoustic, X-ray, multispectral imaging, and other intraoperative imaging modalities, force signals, gaze, voice, and physiological measures—requires specialized hardware integration, consent processes, and seamless coordination within real-world procedures. This must be done without disrupting workflow or compromising patient care, making large-scale data acquisition slow, expensive, and often complicated in many clinical environments.

Second, data sharing is restricted by legal, ethical, and institutional barriers. Hospitals and research centers are often hesitant to share data due to concerns over patient privacy, data ownership, and liability. Even when technically feasible, aligning data formats, standards, and metadata across institutions is nontrivial. Federated learning has emerged as a potential solution, but it has not been very successful in its early stages and introduces new complexities.  Generative AI has opened the path towards synthetic data generation based on limited patient data and clinical knowledge graphs. This is an extremely promising path but it needs further development and focus within the complex multimodal domain of robotic surgery.

Third, annotation is another major challenge. Clinical data requires expert labeling, often by experienced radiologists, surgeons, or sonographers—resources that are both expensive and scarce. Additionally, for dyadic systems, annotations may involve labeling user intent, workflow phases, safety events, or collaboration states—categories that are less well-defined and require task-specific ontologies. Automating or crowdsourcing such annotations is rarely possible due to the expertise required. 

As a result, data scarcity is not just a bottleneck—it is a multi-faceted challenge that limits model generalizability, personalization, and real-world deployment. Addressing this issue will require coordinated efforts across data governance, multi-institutional collaboration, unsupervised or semi-supervised learning methods, and human-in-the-loop annotation tools tailored for the clinical domain.

\subsection{Lack of Standardization and Validation Protocols}
As medical robotics progresses toward intelligent, collaborative systems—particularly those designed for Dyadic Partnership—one of the most pressing gaps is the absence of standardized protocols for development, validation, and benchmarking. Without a common ground for comparing systems, ensuring safety, or quantifying performance in human-robot collaboration, the translation of such technologies into clinical practice remains fragmented and slow~\cite{yip2025robot}.

Currently, no widely accepted platforms or datasets exist to evaluate how well robotic systems interpret human intent, manage shared control, or adapt across varying users and tasks. Each research group often builds custom pipelines, evaluation criteria, and simulation environments, making results difficult to reproduce or compare. This slows progress and undermines confidence in claimed improvements.

For example, in robotic ultrasound, there is currently no publicly available dataset that captures synchronized probe motion, force feedback, anatomical annotations, and user inputs such as gaze or speech—despite the critical role these modalities play in training and validating dyadic systems. In laparoscopic surgical robotics, physical platforms like the da Vinci Research Kit (dVRK) and simulation environments such as Surgical Action Triplet (SAT) provide useful benchmarks, but they offer limited support for evaluating multimodal human-robot interaction and intent recognition. 
MM-OR~\cite{ozsoy2025mm} and EgoExOR~\cite{ozsoy2025egoexor} provide different rich emulated surgical dataset in realistic environments for robotic total knee replacement and multiple other procedures. However, while excellent for holistic understanding of the surgery and generation of intelligent co-pilot which are essential for Dyadic Partnerships, they do not still provide tools and metrics for standardization of provided solutions. Moreover, due to the diversity of clinical applications and task structures, a significant gap exists between datasets for different procedures, making it difficult to directly compare or generalize methods across domains.

To move forward, the field urgently needs shared datasets, open-source platforms, and task-specific benchmarks that capture the complexity of clinical collaboration. These should include:
\begin{enumerate}[1.]
\item Multimodal recordings of real procedures with human annotations for intent, task phases, and errors.
\item Benchmark tasks (e.g., collaborative scanning, shared tool guidance) with metrics for safety, adaptability, and user satisfaction. 
\item Simulation environments that allow reproducible comparisons of control strategies, co-adaptation, and response to uncertainty.
\end{enumerate}

Such platforms would not only support reproducibility but also facilitate regulatory acceptance, accelerate training of foundation models, and align efforts across academia, industry, and healthcare providers. Without these standardization efforts in mid- and long-term, dyadic partnership in medical robotics risks remaining a research concept rather than a clinically validated reality.

\subsection{Regulatory and Social Acceptance}
As intelligent agents move from tools to collaborative partners in clinical care, their ethical and social acceptance becomes essential. In dyadic systems where control is discussed, agreed upon and possibly shared, accountability can become ambiguous—for instance, if a surgical robot assists in a dissection and a complication occurs, it is unclear whether the fault lies with the clinician, the system, or the manufacturer.

Patient autonomy and informed consent are also at stake. A semi-autonomous ultrasound robot, for example, must clearly communicate what it will do, what data it collects, and how much control the patient or clinician retains. Without transparency, users may feel a loss of control or mistrust the system. Bias is another risk. If a robot is trained only on data from a single hospital or population, it may underperform on diverse patients or with clinicians using different techniques. Addressing this requires careful dataset design and fairness evaluation.

Finally, cultural and psychological factors play a critical role in the acceptance of robotic systems in real deployments. In some settings, patients may feel more at ease with robotic guidance—such as in tele-ultrasound scenarios—while in others, increased automation may be perceived as impersonal or intrusive. To gain widespread acceptance, dyadic systems must be thoughtfully designed to align with social values, clinical norms, and human comfort. A compelling example is the recent development of a virtual intelligent sonographer, where an avatar embedded with physiological awareness of the ongoing procedure~\cite{song2025intelligent, song2025enhancing}. This system enables synchronized interaction between the virtual sonographer, the physician or real sonographer, and the patient, offering a more human-centric and emotionally attuned robotic experience. This system is one of the first example of a Dyadic Partnership in which the physician and the virtual sonographer discuss and agree on planned acquisition and data analysis.

\section{Conclusion}
Dyadic Partnership marks a transformative shift in medical robotics—one that moves decisively beyond traditional, tool-like automation toward systems that can reason, adapt, and collaborate alongside clinicians in complex, dynamic environments. By integrating generalizable foundation models, multimodal intent recognition, adaptive control sharing, co-learning mechanisms, and explainable interaction strategies, this framework offers the first comprehensive foundation for truly collaborative human–robot interaction in clinical care. Crucially, Dyadic Partners do not operate under a master–slave hierarchy: they function as equitable collaborators, each contributing complementary strengths grounded in human experience and robotic precision. This paradigm opens a vast new research frontier spanning trust, communication, shared decision-making, and ethical frameworks for next-generation medical systems.

The imperative for Dyadic Partnership becomes especially evident as robotic capabilities begin to surpass human perceptual, cognitive, and physical limits. Future robotic systems may coordinate multiple arms and heterogeneous tools—such as fluoroscopy, laser ablation, ultrasound cutting, or focused ultrasound therapy—executing high-dimensional, multi-parameter actions that no human operator can intuitively monitor, reason about, or directly control in real time. These platforms will not merely automate existing clinical workflows; rather, they will introduce fundamentally new operational paradigms. Representative examples include magnetic-driven microrobotic systems for targeted drug delivery~\cite{landers2025clinically}, magnetic robotic endoscopy~\cite{martin2020enabling}, and non-contact, tissue-selective therapeutic technologies such as Histosonics’ histotripsy system. In such settings, autonomy does not eliminate clinical responsibility. On the contrary, the clinician must remain fully informed, situationally aware, and accountable for high-level decision-making and oversight. This emerging reality highlights the limitations of traditional human-centric control models, which assume that the human operator can directly perceive and command all critical system states. Instead, it calls for a shift toward a true Dyadic Partnership, in which humans and robots exchange information transparently, reason collaboratively, and coordinate actions in a complementary manner—enabling outcomes that neither human expertise nor robotic autonomy could achieve in isolation.

Ultimately, Dyadic Partnership offers a path toward a more capable, equitable, and resilient future of healthcare—one where human judgment and robotic intelligence are not competing forces, but synergistic partners shaping a new era of precision, safety, and accessibility in medicine.

\bibliography{sn-bibliography}

\end{document}